%% file: 0-main.tex
\documentclass[10pt,twocolumn,letterpaper]{article}

\usepackage{cvpr}              

\usepackage{graphicx}
\usepackage{amsmath}
\usepackage{amssymb}
\usepackage{booktabs}

\usepackage[pagebackref,breaklinks,colorlinks]{hyperref}

\input{style}

\def\cvprPaperID{1} 
\def\confName{CVPR}
\def\confYear{2023}

\begin{document}


\title{Are Labels Needed for Incremental Instance Learning?}

\author{Mert Kilickaya\\
Eindhoven University of Technology\\
{\tt\small kilickayamert@gmail.com}
\and
Joaquin Vanschoren\\
Eindhoven University of Technology\\
{\tt\small j.vanschoren@tue.nl}
}
\maketitle

\begin{abstract}

In this paper, we learn to classify visual object instances, incrementally and via self-supervision (self-incremental). Our learner observes a single instance at a time, which is then discarded from the dataset. Incremental instance learning is challenging, since longer learning sessions exacerbate forgetfulness, and labeling instances is cumbersome. We overcome these challenges via three contributions: \textit{i).} We propose VINIL, a self-incremental learner that can learn object instances sequentially, \textit{ii).} We equip VINIL with self-supervision to by-pass the need for instance labelling, \textit{iii).} We compare VINIL to label-supervised variants on two large-scale benchmarks~\cite{core50,ilab20m}, and show that VINIL significantly improves accuracy while reducing forgetfulness.

\end{abstract}

\input{1-introduction}

\input{2-relwork}

\input{3-method}

\input{4-experiments}

\input{5-conclusion}

{\small
\bibliographystyle{ieee_fullname}
\bibliography{egbib}
}

\end{document}

%% file: style.tex
\usepackage{times}
\usepackage{epsfig}
\usepackage{graphicx}
\usepackage{amsmath}
\usepackage{amssymb}

\usepackage{bbm}

\usepackage{amssymb}
\usepackage{caption}
\usepackage{tabularx,ragged2e}
\usepackage{spreadtab}
\usepackage{adjustbox}
\usepackage{booktabs}
\usepackage{xcolor}
\usepackage{multirow}
\usepackage{rotating}
\usepackage{color, colortbl}

\usepackage[inline]{enumitem}
\usepackage{color,soul}

\graphicspath{{./figures/}}

\DeclareGraphicsExtensions{.jpeg,.png,.eps,.pdf}

\definecolor{Gray}{gray}{0.9}

\definecolor{Gray}{gray}{0.85}
\definecolor{LightCyan}{rgb}{0.88,1,1}

\newcolumntype{a}{>{\columncolor{Gray}}c}
\newcolumntype{b}{>{\columncolor{white}}c}

\newcommand{\partitle}[1]{\bigbreak\noindent\textbf{#1}}

\makeatletter
\newcommand*{\rom}[1]{\expandafter\@slowromancap\romannumeral #1@}
\makeatother

\usepackage{mathtools}

\usepackage{pifont}
\newcommand{\xmark}{\ding{55}}%

\makeatletter
\newcommand*\bigcdot{\mathpalette\bigcdot@{.5}}
\newcommand*\bigcdot@[2]{\mathbin{\vcenter{\hbox{\scalebox{#2}{$\m@th#1\bullet$}}}}}
\makeatother

\definecolor{applegreen}{rgb}{0.55,0.71,0.0}
\definecolor{babyblue}{rgb}{0.54,0.81,0.94}
\definecolor{azure}{rgb}{0.0,0.5,1.0}
\definecolor{budgreen}{rgb}{0.48,0.71,0.38}
\definecolor{amaranthpurple}{rgb}{0.67,0.15,0.31}

\usepackage{cleveref}

\crefformat{section}{\S#2#1#3} 
\crefformat{subsection}{\S#2#1#3}
\crefformat{subsubsection}{\S#2#1#3}

\usepackage{booktabs,arydshln}

\makeatletter
\def\adl@drawiv#1#2#3{%
        \hskip.5\tabcolsep
        \xleaders#3{#2.5\@tempdimb #1{1}#2.5\@tempdimb}%
                #2\z@ plus1fil minus1fil\relax
        \hskip.5\tabcolsep}
\newcommand{\cdashlinelr}[1]{%
  \noalign{\vskip\aboverulesep
           \global\let\@dashdrawstore\adl@draw
           \global\let\adl@draw\adl@drawiv}
  \cdashline{#1}
  \noalign{\global\let\adl@draw\@dashdrawstore
           \vskip\belowrulesep}}
\makeatother

\definecolor{LightCyan}{rgb}{0.88,1,1}

%% file: 1-introduction.tex
\section{Introduction}




This paper strives for incrementally learning to recognize visual object instances. Visual instance recognition aims to retrieve different views of an input object instance image. It can be seen as fine-grained object recognition, where the goal is to distinguish different instantiations of the same object, such as cup 1 from cup 2. Instance recognition finds applications in many domains, such as in visual search~\cite{oh2016deep}, tracking~\cite{bertinetto2016fully,shuai2021siammot,tao2016siamese} and localization~\cite{zhu2021vigor}.

Distinguishing between different object instances is a challenging task as they often differ only by small nuances. Metric learning~\cite{wang2017deep} is a commonly used approach to learn visual object instances by comparing two views of the same object using a deep convolutional network, such as ResNet~\cite{resnet}. The network is trained to bring representations of the same object close together and separate representations of different objects in a large batch of images.

However, this approach requires iterating over potentially million-scale datasets multiple times to refine the metric space, which can be impractical for privacy reasons (some data may have to be deleted) or scale (when dealing with billions of images). Additionally, using the trained deep net to query a large database of images by comparing the feature representation of the input image to the database representations is time-consuming and computationally expensive.


This paper builds upon incremental learning to mitigate privacy and scale issues. In incremental learning, the learner observes images from a certain class for a number of iterations. Then, the data of the previous class is discarded, and the learner receives examples from a novel category. Such approach is called class-incremental learning, and receives an increasing amount of attention recently~\cite{masana2020class,kang2022class,mittal2021essentials,zhu2021class,kilickaya2023towards}.  


Existing class-incremental learners are ill-suited for instance-incremental learning for two reasons. First, class-incremental learners rely on full label supervision. Collecting such annotation at the instance level is very expensive. Second, despite years of efforts, class-incremental learners are forgetful, since they lose performance on previously observed categories. 

\begin{figure*}[t]
    \centering
\includegraphics[width=\textwidth]{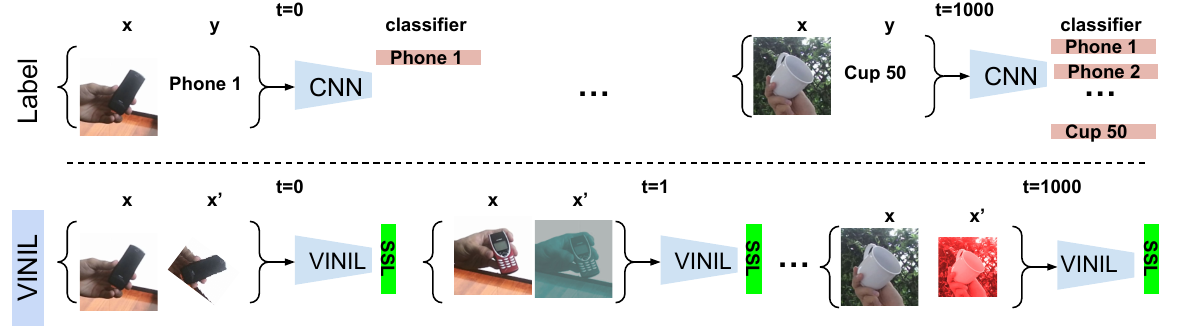}
\caption{\textbf{Top:} Label-incremental learning demands instance-level annotations and trains a new weight per instance. This approach is not suitable for handling large numbers of visual instances, and it is also prone to forgetting previously learned instances. \textbf{Bottom:} In this paper, we introduce VINIL, a self-incremental instance learning method. VINIL focuses solely on learning a discriminative embedding and uses Self-Supervised Learning (SSL) to extract supervision from different views of the same instance. As a result, VINIL is label-free, more scalable, and significantly less prone to forgetting compared to label-incremental learning.}
    \label{fig:teaser}
\end{figure*}


This paper proposes \textbf{V}isual Self-\textbf{In}cremental \textbf{I}nstance \textbf{L}earning, VINIL, to perform instance-incremental learning, consider Figure~\ref{fig:teaser}. VINIL observes multiple views of a single instance at a time, which is then discarded from the dataset. Such examples can be easily captured via turntable cameras~\cite{turntable1,turntable2,turntable3,ilab20m} or via hand-interactions~\cite{handobject1,handobject2,handobject3}. Then, VINIL extracts its own supervision via self-supervision~\cite{barlowtwins}, therefore self-incremental. Self-incremental learning not only is label-efficient, it also consistently outperforms competitive label-supervised variants, as we will show. In summary, this paper makes three contributions:

\begin{enumerate}[label=\Roman*.]




\item In this paper, we study the challenging task of incremental visual instance learning, 
\item We propose VINIL, an incremental instance learner solely guided by self-supervision, by-passing the need for heavy supervision,
\item Through large-scale experiments on~\cite{core50,ilab20m}, we show that VINIL is more accurate and much less forgetful with respect to competitive label-supervised variants, hence unlocking the potential of large-scale incremental learning for free. 



\end{enumerate}

%% file: 2-relwork.tex
\section{Related Work}

\partitle{Visual Instance Recognition.} Visual instance recognition has been extensively researched in recent years and has been applied to various computer vision problems, including product retrieval~\cite{oh2016deep,hermans2017defense,liu2017sphereface,wang2017deep}, object tracking~\cite{bertinetto2016fully,shuai2021siammot,tao2016siamese}, and geo-localization~\cite{lin2013cross,vyas2022gama,zhu2022transgeo,shi2022beyond,wilson2021visual}. The most common approach for these tasks is to induce a discriminative embedding space, often using metric learning techniques~\cite{hermans2017defense,chopra2005learning}. These methods require access to the entire dataset and fine-grained similarity labels. In contrast, this paper presents a novel method for incremental and label-free visual instance recognition, in a similar vein to~\cite{parshotam2020continual}.


\partitle{Class-Incremental Learning.} Class-incremental learning involves expanding a deep classifier with novel objects, with the goal of maintaining performance on previous categories and avoiding forgetting~\cite{masana2020class}. Popular techniques to prevent forgetting include regularization, which limits abrupt changes in network weights~\cite{kirkpatrick2017overcoming,li2017learning,riemer2018learning,kerssies2022evaluating}, and memory replay of previous data~\cite{balaji2020effectiveness,shin2017continual,rolnick2019experience,ho2021prototypes}. Our approach differs from conventional class-incremental learning in two ways. First, while class-incremental learning focuses on object categories, our approach operates at the instance level, presenting new challenges. Second, class-incremental learning requires fully labeled datasets, which is often not possible in instance learning. To overcome these limitations, we use self-supervision and adapt relevant evaluation techniques. Specifically, we use Elastic Weight Consolidation (EwC) as a regularization method~\cite{kirkpatrick2017overcoming} and Replay as a memory technique~\cite{rolnick2019experience} due to their adaptability for label-free learning.

\partitle{Self-Supervised Learning.} Self-supervision involves creating pretext tasks to learn deep representations without using labels. Early methods predicted rotations~\cite{gidaris2018unsupervised} or patches~\cite{noroozi2016unsupervised}, but contrastive learning has become dominant in recent years~\cite{chen2020improved,caron2020unsupervised,chen2020simple,he2020momentum}. In our work, we utilize self-supervision to extract learning signals in place of instance labels. We experimented with both BarlowTwins~\cite{barlowtwins} and SimSiam~\cite{simsiam} due to their high performance and adaptation in incremental learning tasks~\cite{fini2022self,madaan2021representational}. We found that BarlowTwins~\cite{barlowtwins} performs better than SimSiam for our incremental learning setup. We believe this is due to its ability to reduce redundancy across different views of the input. Reducing visual redundancy is especially important for different instances of the same object, as visual object instances may only differ in small details.

\partitle{Incremental Self-Supervised Learning.} Recently, there has been a surge of interest in use of self-supervision to replace label supervision for incremental learning. We identify three main directions. \textit{i) Pre-training: } Researchers use self-supervised learning either for pre-training prior to incremental learning stage~\cite{gallardo2021self,caccia2021special,hu2021well} or as an auxiliary loss function to improve feature discrimination~\cite{zhu2021prototype}. However, these papers still require labels during the incremental learning stage. \textit{ii) Replay: } Second line of techniques propose replay-based methods~\cite{purushwalkam2022challenges,cha2021co2l,madaan2021representational} to supplement self-supervised learners with stored data within the memory. \textit{iii) Regularization: } Third line of work proposes to regularize self-learned representations~\cite{fini2022self,gomez2022continually,madaan2021representational}.  

In this work, we focus on replay and regularization-based self-incremental learning. More specifically, we closely follow UCL~\cite{madaan2021representational} and ask ourselves: What is the contribution of self-supervision for instance incremental learning? Our main observation is that self-supervision consistently yields less forgetful, more accurate and transferable representations, as will be shown via large-scale experiments. 



%% file: 3-method.tex
\begin{table*}[t]
\begin{center}
\begin{tabular}{lcccc}

\toprule 

Method  & Supervision  & Input & Memory & Loss \\ 
\midrule 
\midrule 
 Fine-Tuning    & Label-supervised & $(x, y)$  & \xmark & $CE(y, y^\prime)$ \\ 
\rowcolor{LightCyan} Fine-Tuning    & Self-supervised & $(x)$  & \xmark & $BT(x, x^{\prime})$ \\

 \cmidrule{1-5}
 
  EwC    & Label-supervised & $(x, y)$  & \xmark & $CE(y, y^\prime) + Reg(\Theta, y^{\prime}) $ \\ 
\rowcolor{LightCyan}  EwC    & Self-supervised & $(x)$  & \xmark & $BT(x, x^{\prime}) + Reg(\Theta)$ \\ 

 \cmidrule{1-5}
 Replay    & Label-supervised & $(x, y)$  & $(x^{m}, y^{m})$ & $CE(y, y^\prime) + CE(y^{m}, y^{m^{\prime}})$ \\ 
\rowcolor{LightCyan} Replay    & Self-supervised & $(x)$  &  $(x^{m})$ & $BT(x, x^{\prime}) + BT(x^{m}, x^{m^{\prime}})$ \\

\bottomrule
\end{tabular} 
\end{center}

\caption{VINIL performs incremental instance learning via self-supervision, and is compared with label-supervision. We use memory replay~\cite{rolnick2019experience} and weight regularization~\cite{kirkpatrick2017overcoming} as well as simple fine-tuning. \textit{Fine-Tuning}~\cite{sgd} relies on Cross-Entropy \textbf{(CE)} or BarlowTwins \textbf{(BT)}~\cite{barlowtwins} to perform incremental learning. \textit{EwC}~\cite{kirkpatrick2017overcoming} penalizes abrupt changes in network weights via regularization ($Reg(\cdot)$). \textit{Replay}~\cite{rolnick2019experience} replays a part of previous data in the form of input-labels (label-supervised) or input-only (self-supervised).} 

\label{tab:method}
\end{table*}

\section{VINIL}
We present an overview of VINIL in Table~\ref{tab:method}. The goal of VINIL is to train an embedding network $f_{\theta_{t}}(\cdot)$ parameterized by $\theta_{t}$. The network maps an input image $x$ to a $D$-dimensional discriminative embedding, $h = f_{\theta_{t}}(x)$ which will then be used to query the database to retrieve different views of the input query for instance recognition. Here, $t$ denotes the incremental learning step, where the tasks are arriving sequentially: $\mathbf{T} = (\mathbf{T}_{1}, \mathbf{T}_{2}, ..., \mathbf{T}_{t})$. We train VINIL via minimizing the following objective: 
\begin{align}
    \mathcal{L} = w_{c} \cdot L_{inst} + (1 - w_{c}) \cdot L_{incr}
\end{align}
\noindent where $w_{c}$ controls the contribution of instance classification loss $L_{inst}$ and incremental learning loss $L_{incr}$. Incremental learning loss either corresponds to memory replay~\cite{rolnick2019experience} or weight regularization~\cite{kirkpatrick2017overcoming} whereas instance classification loss $L_{inst}$ is either cross-entropy with labels or a self-supervision objective. 

\subsection{Incremental Learning}

\partitle{Fine-Tuning (FT).} A vanilla way to perform incremental instance learning is to apply simple fine-tuning via SGD~\cite{sgd}. In fine-tuning, no incremental learning loss is applied (\ie $w_{c} = 1.0$) and the sole objective is classification. 

In case of label-supervision, a task is defined by a dataset $D_{t}^{label} = \{ (x_{i, t}, y_{i, t})_{i=1}^{k_{t}} \}$ where $k_{t}$ is the data size at time $t$. Then, fine-tuning corresponds to instance discrimination via cross-entropy $L_{inst} = CE (y_{i, t}, y_{i, t}^{\prime})$. Here, instance category prediction for the instance $i$ at time step $t$ is obtained with a simple MLP classifier. Notice that this classifier will expand in size linearly with the number of instance categories. 

In case of VINIL, a task is defined by a dataset $D_{t}^{self} = \{ (x_{i, t})_{i=1}^{n_{t}} \}$ (\ie no labels). Then, fine-tuning corresponds to minimizing the self-supervision objective $L_{inst} = BT (x_{i, t}, x_{i, t}^{\prime})$ where $BT(\cdot)$ is the BarlowTwins~\cite{barlowtwins}.

\partitle{EwC~\cite{kirkpatrick2017overcoming}.} EwC penalizes big changes in network weights via comparing the weights in the current and the previous incremental learning step. Originally, EwC re-weights the contribution of each weight to the loss function as a function of instance classification logits (\ie label-supervision). In VINIL, in the absence of labels, we omit this re-weighting and simply use identity matrix.  

\partitle{Replay~\cite{rolnick2019experience}.} Replay replays a portion of the past data from previous incremental steps to mitigate forgetting. In case of label-supervision, this corresponds to replaying both the input data and their labels via cross-entropy: $ CE(y_{i, t}^{m}, y_{i, t}^{m^{\prime}})$ where $y_{i, t}^{m^{\prime}}$ is the instance categories for the memory instance $i$ at time $t$. For VINIL, we simply replay the input memory data and its augmented view via self-supervision of BarlowTwins as $BT(x_{i, t}^{m}, x_{i, t}^{m^{\prime}})$. 


\subsection{Self-Supervised Learning}

In BarlowTwins, the features are extracted from the original and the augmented view of the input image with a siamese deep network, at time step $t$ as: $(z_{i, t}, z_{i, t}^{\prime}) = (f_{\theta^{t}}(x_{i, t}), f_{\theta^{t}}(x_{i, t}^{\prime}))$ where $x_{i, t}^{\prime} = aug(x{i, t})$ is the augmented view of the input. BarlowTwins minimizes the redundancy across views while maximizing the representational information. This is achieved via operating on the cross-covariance matrix via:  

\begin{align}
BT = \sum_{i}(1 - C_{ii})^2 + w_{b} \cdot \sum_{i} \sum_{j \neq i}(C_{ij})^2
\end{align}

\noindent where:

\begin{align}
  C_{ij} = \frac{\sum_{\beta} z_{\beta, i} z^{\prime}_{\beta, j}}{\sum_{\beta} \sqrt{z_{\beta, i}^2} \cdot \sum_{\beta} \sqrt{(z_{\beta, j}^{\prime})^2}}
\end{align}

\noindent is the cross-correlation matrix. Here, $w_{b}$ controls invariance-redundancy reduction trade-off, $i$ and $j$ corresponds to network's output dimensions.

%% file: 4-experiments.tex
\section{Experimental Setup}

\partitle{Implementation.} All the networks are implemented in PyTorch~\cite{pytorch}. We use ResNet-$18$~\cite{resnet} as the backbone $f(\cdot)$, and a single-layer MLP for the instance classifier. We train for $200$ epochs for each incremental steps with a learning rate $0.001$ decayed via cosine annealing. We use SGD optimizer with momentum $0.9$ and batch-size $256$. We use random cropping and scaling for augmentation. 

We follow the original implementation of BarlowTwins~\cite{BarlowTwins_code}. $10\%$ of the data is stored within the memory for replay~\cite{rolnick2019experience}. We set $w_{c} = 0.7$ and $w_{b} = 0.03$.

\partitle{Datasets.} We evaluate VINIL on iLab-$20$M~\cite{ilab20m} and Core-$50$~\cite{core50}, since they are large-scale, sufficiently different, and widely adopted in incremental learning.  

\textit{iLab-$20$M} is a turntable dataset of vehicles. It consists of $10$ objects (\ie bus, car, plane) with varying ($[25,160]$) number of instances per category. Objects are captured by varying the background and the camera angle, leading to $14$ examples per-instance. We use the public splits provided in~\cite{iLab_data_split} with $125$k training and $31$k gallery images. 

\textit{Core-$50$} is a hand-held object dataset used in benchmarking incremental learning algorithms. The dataset includes $10$ objects (\ie phones, adaptors, scissors) with $50$ instances per-category. Each instance is captured for $300$ frames, across $11$ different backgrounds. We use $120$k training and $45$k gallery images~\cite{core50_data_split}.


\partitle{Protocol.} We first divide each dataset into $5$ tasks, with $2$ object categories per-task. Then, each task is subdivided into $N$ object instance tasks depending on the dataset. We discard the classifier of label-supervised variants after training, and evaluate all models with instance retrieval performance via k-NN with $k=100$ neighbors on the gallery set, as is the standard in SSL~\cite{chen2020improved,caron2020unsupervised,chen2020simple,he2020momentum, simsiam}. 

We use the mean-pooled activations of \textsc{layer4} of ResNet to represent images. All exemplars in the gallery set are used as query.


\partitle{Metrics.} We rely on two well established metrics to evaluate the performance of the models, namely accuracy and forgetting. 

\textit{i). Accuracy} (Acc) measures whether if we can retrieve different views of the same instance from the gallery set given a query. We measure accuracy for each incremental learning steps, which is then averaged across all sessions.

 \textit{ii). Forgetting} (For) measures the discrepancy of accuracy across different sessions. Concretely, it compares the maximum accuracy across all sessions with the accuracy in the last step.




\section{Experiments}


Our experiments address the following research questions: \textbf{Q1}: Can VINIL improve performance and reduce forgetting in comparison to label-supervision? \textbf{Q2}: Does VINIL learn incrementally generalizable representations across datasets? \textbf{Q3}: What makes VINIL effective against label-supervision? 

\subsection{How Does VINIL Compare to Label-Supervision?}

First, we compare VINIL's performance to label-supervision. The results are presented in Table~\ref{tab:category}.

\begin{table}[h]
    \centering
    \setlength{\tabcolsep}{5pt}
    \begin{tabular}{l c  c c c}
        \toprule
         & \multicolumn{2}{c}{Core-$50$} & \multicolumn{2}{c}{iLab-$20$M} \\
        \cmidrule(lr){2-3} \cmidrule(lr){4-5}
       Method &  $Acc$ ($\uparrow$) & $For$ ($\downarrow$) & $Acc$ ($\uparrow$) & $For$ ($\downarrow$) \\
        \midrule
        FT (Label) & 71.450 & 22.436 & 89.340 & 6.500 \\
        \rowcolor{LightCyan} FT (VINIL) & \textbf{74.914} & \textbf{4.802} & \textbf{90.398} & \textbf{0.000} \\
        \midrule
        Replay (Label) & \textbf{88.180} & \textbf{6.741} & 84.464 & 5.696 \\
        \rowcolor{LightCyan} Replay (VINIL) & 67.677 & 10.095 & \textbf{90.543} & \textbf{0.000} \\
        \midrule
        EwC (Label) & \textbf{75.117} & 18.268 & 87.690 & 4.535 \\
        \rowcolor{LightCyan} EwC (VINIL) & 73.011 & \textbf{2.167} & \textbf{90.655} & \textbf{0.000} \\
        \bottomrule
    \end{tabular}
    \caption{Visual Incremental Instance Learning on Core-$50$~\cite{core50} and iLab-$20$M~\cite{ilab20m}. VINIL outperforms label-supervised variants for $4$ out of $6$ settings, while significantly reducing forgetfulness on both datasets. This indicates self-incremental learning is a strong, label-free alternative to label-supervision.}
    \label{tab:category}
\end{table}

\partitle{VINIL Yields Competitive Accuracy.} We first compare the accuracies obtained by VINIL vs. label-supervision. We observe that VINIL yields competitive accuracy against label-supervision: In $4$ out of $6$ setting, VINIL outperforms label-supervised variants. 



\partitle{VINIL Mitigates Forgetting.} Secondly, we compare the forget rates of VINIL vs. label-supervision (lower is better). We observe that VINIL consistently leads to much lower forget rates in comparison to label-supervision. On iLab-$20$M dataset, VINIL results in \textit{no} forgetting. On the more challenging dataset of Core-$50$, the difference across forget rates are even more pronounced: Label-supervision suffers from $22\%$ forget rate whereas VINIL only by $4\%$, a relative drop of $80\%$ with fine-tuning. 



\partitle{Label-supervision Leverages Memory.} Our last observation is that memory improves the accuracy and reduces forgetfulness of label-supervision. In contrast, the use of memory disrupts self-supervised representations. This indicates that replaying both inputs and labels ($(x_{i}, y_{i})$) as opposed to input-only ($(x_{i})$, as in self-supervision) may lead to imbalanced training due to limited memory size~\cite{Castro2018EndtoEndIL,Hou2019LearningAU,Wu2019LargeSI}. 





In summary, we conclude that VINIL is an efficient, label-free alternative to label-supervised incremental instance learning. VINIL improves accuracy while reducing forget rate. We also observe that label-supervision closes the gap when an additional memory of past data is present. This motivates further research for improving self-incremental instance learners with memory.

\subsection{Can VINIL Generalize Across Datasets?}

After confirming the efficacy of VINIL within the same dataset, we now move on to a more complicated setting: Cross-dataset generalization. In cross-dataset generalization, we first perform incremental training on Core-$50$, and then evaluate on iLab-$20$M. Then, we perform incremental training on iLab-$20$M and then evaluate on Core-$50$. 

Cross-dataset generalization between Core-$50$ and iLab-$20$M is challenging due to the following reasons:  \textit{i). Camera: } Core-$50$ is captured with a hand-held camera whereas iLab-$20$M is captured on a platform with  a turntable camera, \textit{ii). Object Categories:} Object categories are disjoint, as no common objects are present in each dataset, \textit{iii). Object Types:} iLab-$20$M exhibits toy objects of vehicles whereas Core-$50$ exhibits hand-interacted daily-life objects. 


 The results are presented in Table~\ref{tab:cross}. We present iLab-$20$M to Core-$50$, and Core-$50$ to iLab-$20$M results, along with the relative drop w.r.t training and testing on the same dataset (see Table~\ref{tab:category}).

 \begin{table}[h]
\begin{center}
\begin{tabular}{lcc}

\toprule 

 Train on  $\Longrightarrow$ &  iLab-$20$M &    Core-$50$ \\ 

\cmidrule{2-3}


  Test  on  $\Longrightarrow$  & Core-$50$ &      iLab-$20$M   \\ 
\cmidrule{2-3}
Method &    $Acc_{(\%\Delta(\downarrow))}$    & $Acc_{(\%\Delta(\downarrow))}$   \\ 
\midrule 
FT (Label) & $59.850_{\%16}$  & $67.249_{\%24}$   \\ 
\rowcolor{LightCyan} FT (VINIL) &  $\textbf{66.704}_{\textbf{\%10}}$  &  $\textbf{76.302}_{\textbf{\%15}}$  \\ 
\midrule 
Replay (Label) &  $55.692_{\%36}$ & $69.412_{\%17}$  \\ 

\rowcolor{LightCyan} Replay (VINIL) &  $\textbf{61.857}_{\textbf{\%8}}$  &  $\textbf{76.125}_{\textbf{\%15}}$  \\ 

\midrule 
EwC (Label) & $59.030_{\%21}$ &   $70.087_{\%20}$   \\ 


\rowcolor{LightCyan} EwC (VINIL) & $\textbf{70.648}_{\textbf{\%3}}$  & $\textbf{75.793}_{\textbf{\%16}}$  \\ 

\bottomrule
\end{tabular} 
\end{center}
   \caption{Cross-Dataset Generalization on Core-$50$ and iLab-$20$M. We present: \textit{i)} Train on iLab-$20$M and test on Core-$50$, \textit{ii)} Train on Core-$50$ and test on iLab-$20$M. In adddition to accuracy, we also present the relative drop w.r.t training and testing on the same dataset (see Table~\ref{tab:category}). We observe that VINIL is consistently more robust in cross-dataset generalization when compared with label-supervision. The results indicate that self-supervision is able to extract more domain-agnostic representations, which improves the generality of visual representations, for instance-incremental setup.}
\label{tab:cross}
\end{table}

\partitle{VINIL Yields Generalizable Representations.} We first observe that VINIL consistently yields higher accuracy and lower drop rate across all $6$ settings in both datasets. This indicates that self-supervision extracts more generalizable visual representations from the dataset.  



\partitle{Label-supervision Overfits with Memory.} Secondly, we observe that label-supervised variants with memory generalizes via overfitting on the training dataset. Replay with label-supervision leads to the biggest drop rate of $36\%$ on Core-$50$, when trained with iLab-$20$M. This implies the use of the memory drastically reduces generality of visual representations. A potential explanation is that, since replay utilizes the same set of examples within the limited memory repeatedly throughout learning, this forces the network to over-fit to those examples.

We conclude that VINIL extracts generalizable visual representations from the training source to perform instance incremental training. We also conclude that the astounding performance of label-supervision equipped with memory comes with the cost of overfit, leading to drastic drop in case of visual discrepancies across datasets.  


\subsection{What Factors Affect VINIL's Performance?}



\begin{figure}[!tbp]
  \centering
  \begin{minipage}[t]{0.49\textwidth}
    \includegraphics[width=\textwidth]{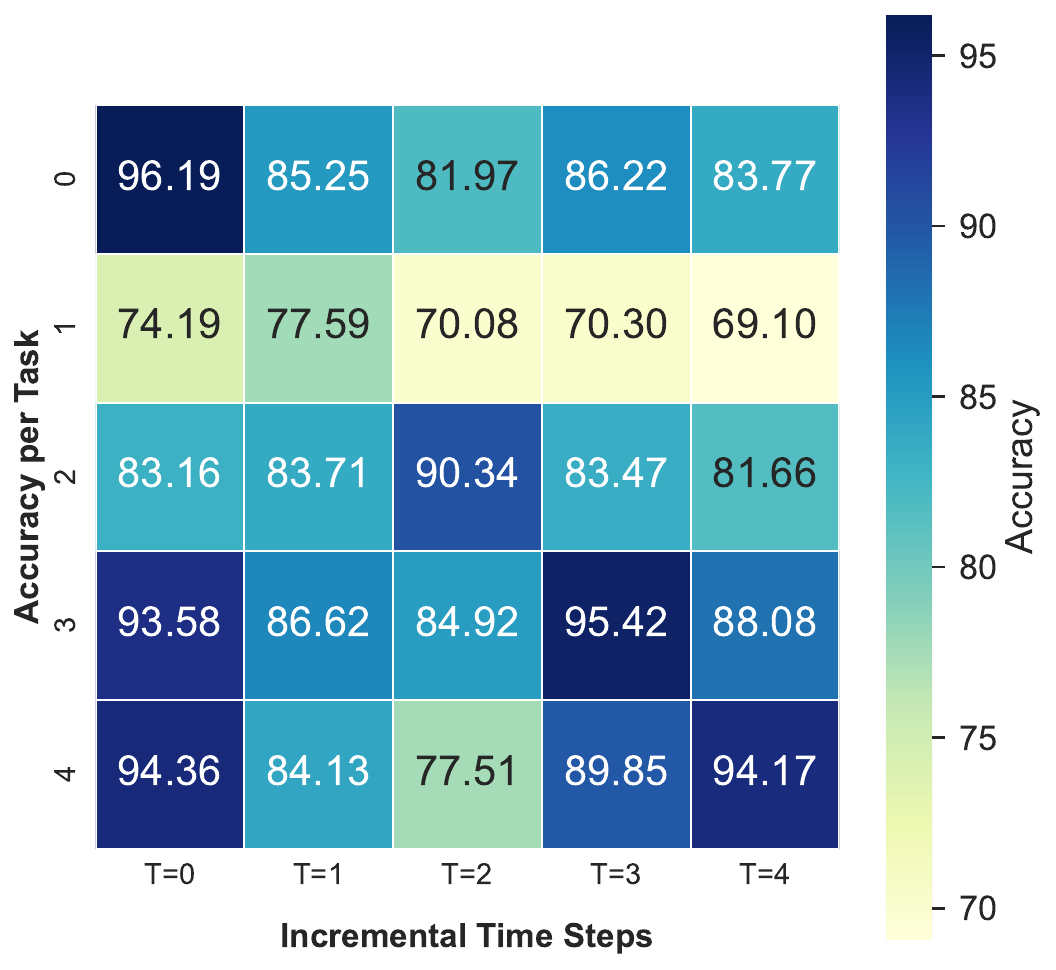}
\caption{Task-level performance of Label-supervision (Fine-tuning). Label-supervision is biased towards recent task.}
    \label{fig:stability1}
  \end{minipage}
  \hfill
  \begin{minipage}[t]{0.49\textwidth}
    \includegraphics[width=\textwidth]{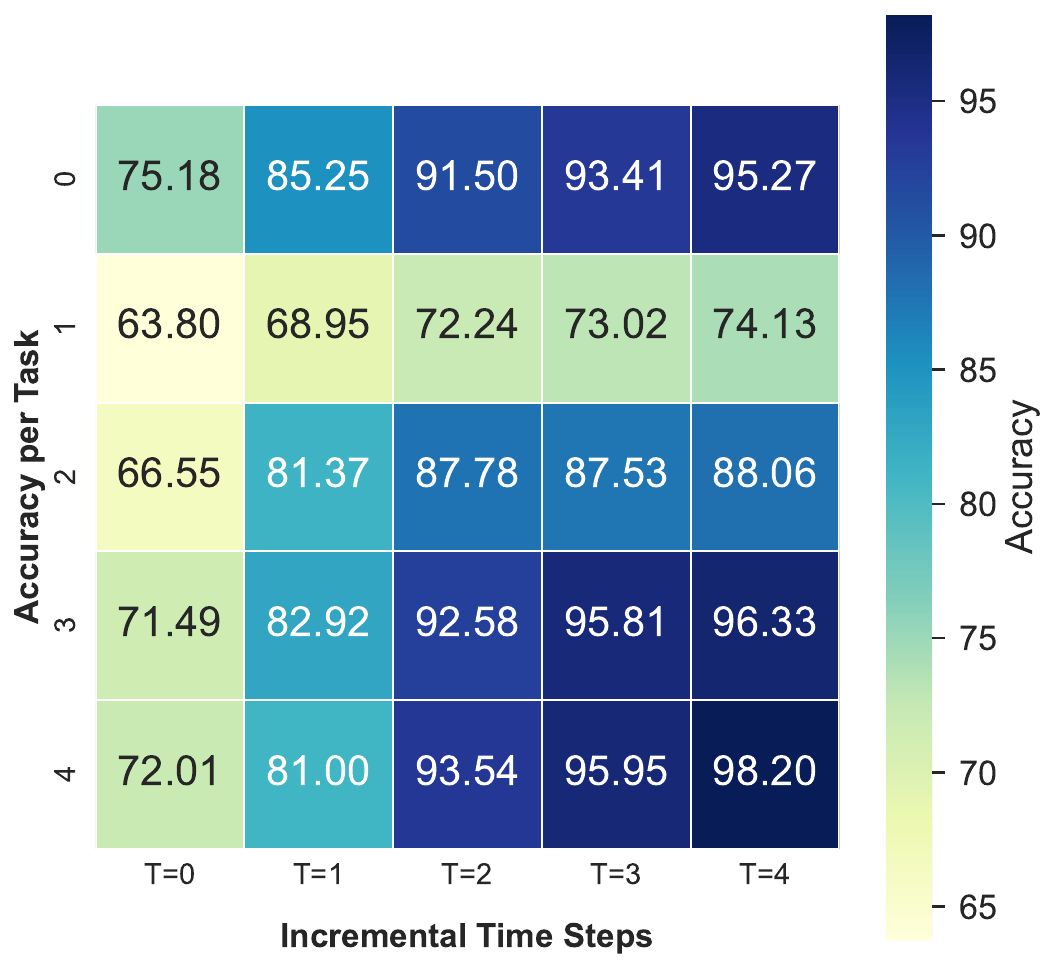}
    \caption{Task-level performance of VINIL (Fine-tuning). VINIL improves its performance with incoming data, and is less biased towards recent task.}
        \label{fig:stability2}
  \end{minipage}

\end{figure}

\partitle{VINIL Mitigates Bias Towards Recent Task.} We present the heatmaps of the performance for all $5$ main tasks, when each task is introduced sequentially, for label-supervision in Figure~\ref{fig:stability1} and for VINIL in Figure~\ref{fig:stability2} on iLab-$20$M~\cite{ilab20m}. Each row presents the accuracy for each task, as the tasks are introduced sequentially. For example, the entry $(0, 2)$ denotes the performance on Task-$0$ when the Task-$2$ is introduced. 


Considering Figure~\ref{fig:stability1} for label-supervision, observe how the tasks achieve their peak performance when they are being introduced to the model, hence the higher numbers within the diagonal. Then, the performance degrades drastically as more and more tasks are being introduced. This indicates label-supervision fails to leverage more data. We call such phenomenon "recency bias", as the model is biased towards the most recently introduced task. 


In contrast, in Figure~\ref{fig:stability2} for VINIL, the performance on each task improves sequentially with the incoming stream of new tasks. This indicates self-supervised representations are less biased towards the recent task, and can leverage data to improve performance. This renders them as a viable option when incremental learning for longer learning steps, such as in incremental instance learning.

\begin{figure}[h]
    \centering
\includegraphics[width=0.45\textwidth]{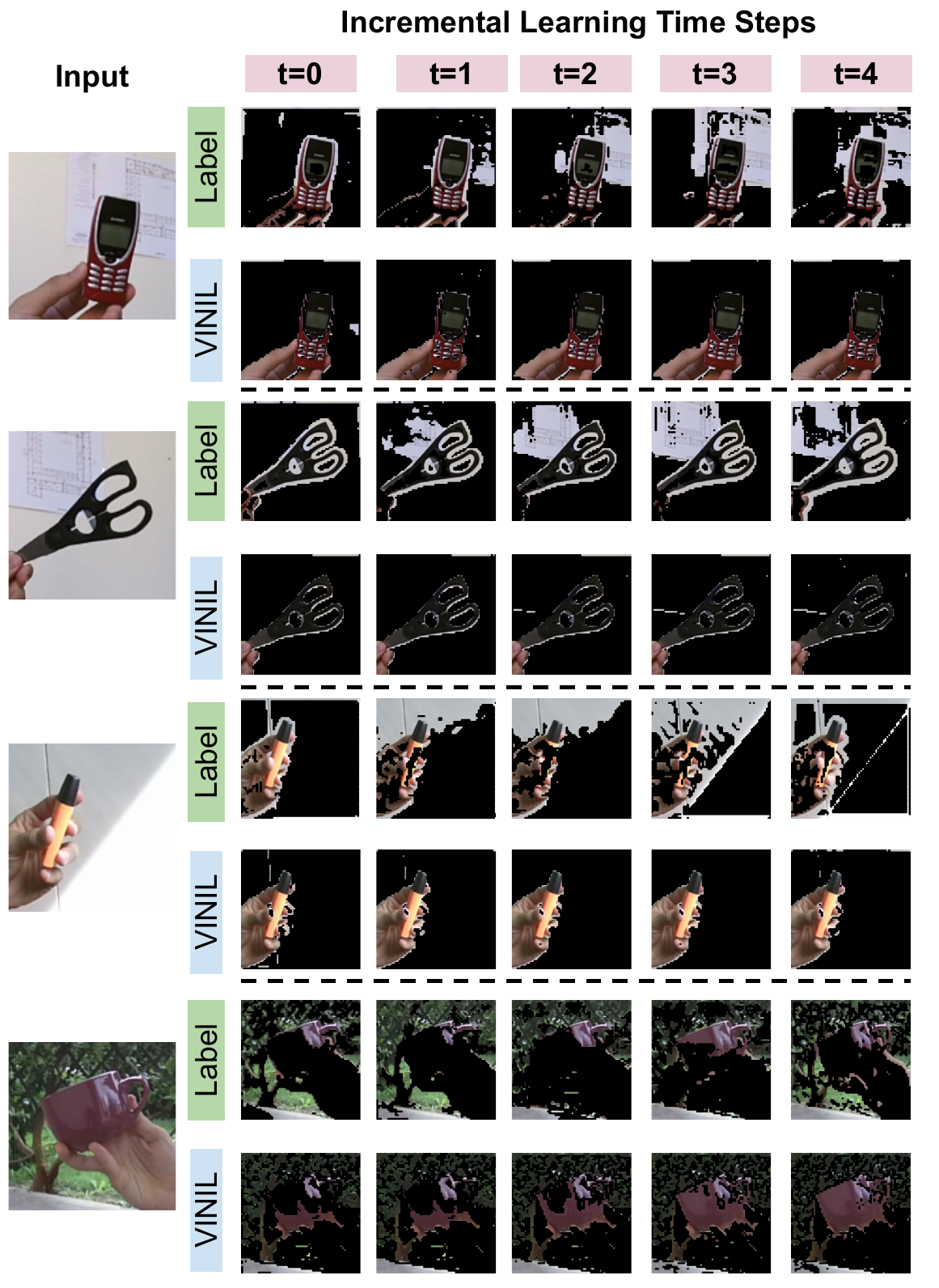}

 \caption{Activations of the last layer of ResNet~\cite{resnet}, throughout the incremental learning steps. We compare label-supervision with VINIL (Fine-tuning). Notice how the attention of the label-supervised variant is disrupted after a few learning tasks. Instead, VINIL learns to segment out the target object, successfully suppressing the background context, such as the hand or the background.}

    \label{fig:activations}
\end{figure}

\partitle{VINIL Focuses on the Object Instance.} We present the activations of the last layer of ResNet, at different incremental time steps, in Figure~\ref{fig:activations}. 

Observe how VINIL learns to segment out the target object from the background. This allows the model to accurately distinguish across different instances of the same object sharing identical backgrounds. In contrast, label-supervised variant progressively confuses the object with the background. We call such a phenomenon "attentional deficiency" of label-supervised representations. 





\begin{figure*}[t]
    \centering
\includegraphics[width=0.85\textwidth]{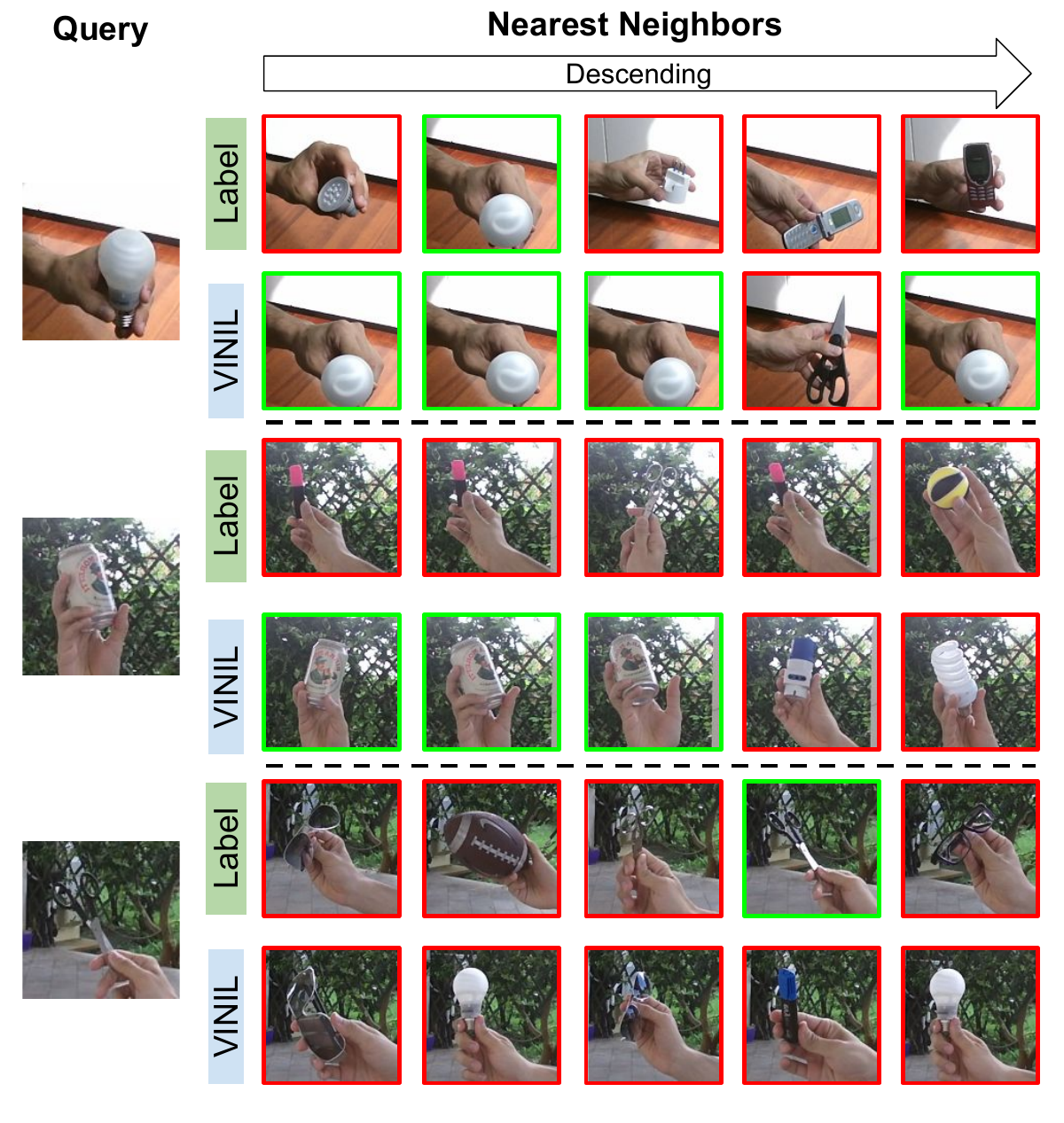}
 \caption{Five nearest neighbors for three object instance queries on Core-$50$~\cite{core50} with fine-tuning. Green is a success, red is a failure. Observe how VINIL retrieves object instances in different views. The last column showcases a failure case, where both models fail to represent an object with holes (scissor).}
    \label{fig:neighbor}
\end{figure*}

\partitle{VINIL Stores Instance-level Information.} We present nearest neighbors for three queries in Figure~\ref{fig:neighbor}. We use the average-pooled activations of the last ResNet layer on Core-$50$ trained with fine-tuning. 

Observe how VINIL retrieves the same instance in different viewpoints, such as for the light bulb and can. In contrast, label-supervision is distracted by the background context, as it retrieves irrelevant objects with identical background. This indicates self-supervision generalizes via storing instance-level information. We present a failure case in the last row, as both models fail to represent an object with holes and un-familiar rotation.



We conclude that VINIL can improve its performance with incoming stream of data, and generalizes via focusing on the target object and storing  instance-level details to perform instance-incremental learning.

%% file: 5-conclusion.tex
\section{Discussion}

This paper presented VINIL, a self-incremental visual instance learner. VINIL sequentially learns visual object instances, with no label supervision, via only self-supervision of BarlowTwins~\cite{barlowtwins}. Below, we summarize our main discussion points:


\textbf{Self \textit{vs}. Label-supervision?} We demonstrate that self-supervision not only omits the need for labels, but it is also more accurate and less forgetful.  

\textbf{W/ or W/o Memory?} Our results show that the use of memory boosts label-supervised instance incremental learning, however the improvement comes with the cost of over-fitting on the training source. 

\textbf{Fine-tuning~\cite{sgd} \textit{vs.} Replay~\cite{rolnick2019experience} \textit{vs.} EwC~\cite{kirkpatrick2017overcoming}?} We demonstrate that with the use of self-supervision, VINIL closes the gap between simple fine-tuning via SGD and more complicated, compute-intensive techniques like memory replay or regularization via EwC. 

\textbf{What Makes VINIL Effective?} VINIL retains representations across tasks, and is able to store and focus on instance-level information, which are crucial for instance-incremental learning. 

\textbf{Limitation.} VINIL is executed with regularization~\cite{kirkpatrick2017overcoming} and memory~\cite{rolnick2019experience}. One can also consider dynamic networks~\cite{Yoon2018LifelongLW} whose architectures are updated with incoming task data. VINIL is a scalable alternative to dynamic incremental network training due to abundant unlabeled data. 

\section{Acknowledgements}
Mert Kilickaya's research is fully funded by ASM Pacific Technology (ASMPT).